\def\year{2019}\relax
\documentclass[letterpaper]{article} 
\usepackage{aaai19}  
\usepackage{times}  
\usepackage{helvet}  
\usepackage{courier}  
\usepackage{url}  
\usepackage{graphicx}  
\usepackage{appendix}
\usepackage{color,threeparttable}
\usepackage{amsmath,amssymb,amsfonts}
\usepackage{algorithm,algorithmic}
\usepackage[english]{babel}
\usepackage{mathrsfs}

\begin{document}
	%
	\title{Direct Training for Spiking Neural Networks:\\ Faster, Larger, Better}
	\author{
		Yujie Wu $^{1}$, Lei Deng $^{2}$, Guoqi Li $^{1}$, Jun Zhu $^{3 \dag}$,Yuan Xie$^{2}$ and Luping Shi $^{1 \dag}$   
		\AND \\
		$^{1}$Center for Brain-Inspired Computing Research, 	Department of Precision Instrument, Tsinghua University \\
		$^{2}$ Department of Electrical and Computer Engineering, University of California, Santa Barbara\\
		$^{3}$ Department of Computer Science and Technology, Institute for AI, THBI Lab, Tsinghua University \\	
	 	 $^{\dag}$Corresponding:   lpshi@tsinghua.edu.cn; dcszj@mail.tsinghua.edu.cn
	}
	\maketitle
	\begin{abstract}
		Spiking neural networks (SNNs)  that enables energy efficient implementation on emerging neuromorphic hardware are gaining more attention. Yet now, SNNs have not shown competitive performance compared with artificial neural networks (ANNs), due to the lack of effective learning algorithms and efficient programming frameworks. We address this issue from two aspects: (1) We propose a neuron normalization technique to adjust the neural selectivity and develop a direct learning algorithm for deep SNNs. (2)  Via narrowing the rate coding window and converting the leaky integrate-and-fire (LIF) model into an explicitly iterative version, we present a Pytorch-based implementation method towards the training of large-scale SNNs. In this way, we are able to train deep SNNs with tens of times speedup. As a result, we achieve significantly better accuracy than the reported works on neuromorphic datasets (N-MNIST and DVS-CIFAR10), and comparable accuracy as existing ANNs and pre-trained SNNs on non-spiking datasets (CIFAR10).  {To our best knowledge, this is the first work that demonstrates direct training of deep SNNs with high performance on CIFAR10, and the efficient implementation provides a new way to explore the potential of SNNs}.
		
	\end{abstract}
	\section{Introduction}
	Spiking neural networks, a sub-category of brain-inspired computing models, use spatio-temporal dynamics to mimic neural behaviors and binary spike signals to communicate between units. Benefit from the event-driven processing paradigm (computation occurs only when the unit receives spike event), SNNs can be efficiently implemented on specialized neuromorphic hardware for power-efficient processing, such as SpiNNaker \cite{Khan2008SpiNNaker}, TrueNorth \cite{Merolla2014Artificial}, and Loihi \cite{Davies2018Loihi}.
	
	As well known, the powerful error backpropagation (BP) algorithm and larger and larger model size enabled by ANN-oriented programming frameworks (e.g. Tensorflow, Pytorch) has boosted the wide applications of ANNs in recent years. However, the rich spatio-temporal dynamics and event-driven paradigm   make SNNs much different from conventional ANNs. Till now SNNs have not yet demonstrated comparable performance to ANNs due to the lack of effective learning algorithms and efficient programming frameworks, which greatly limit the network scale and application spectrum \cite{Tavanaei2018Deep}.  
	
	In terms of learning algorithms, three challenges exist for training large-scale SNNs. First, the complex neural dynamics in both spatial and temporal domains make BP obscure. Specifically, the neural activities not only propagate layer by layer in spatial domain, but also affect the states along the temporal direction, which is more complicated than typical ANNs. Second, the event-driven spiking activity is discrete and non-differentiable, which impedes the BP implementation based on gradient descent. Third, SNNs are more sensitive to parameter configuration because of binary spike representation. Especially in the training phase, we should simultaneously ensure timely response for presynaptic stimulus and avoid too many spikes that will probably degrade the neuronal selectivity. Although previous work has proposed many techniques to adjust firing rate, such as model-based normalization \cite{Diehl2015Fast} and spike-based normalization \cite{Sengupta2018Going}, all of them are specialized for ANNs-to-SNNs conversion learning, not for the direct training of SNNs as considered in this paper. 
	
	In terms of programming frameworks, we lack suitable platforms to support the training of deep SNNs. Although there exist several platforms serving for simulating biological features of SNNs with varied abstraction levels \cite{Brette2007Simulation,Carnevale2006The,Hazan2018BindsNET}, little work is designed for training deep SNNs. Researchers have to build application-oriented models from scratch and the training speed is usually slow. On the other hand, emerging ANN-oriented frameworks   can provide much better efficiency, especially for large models, hence a natural idea is to map the SNN training onto these frameworks. However, these frameworks are designed for aforementioned ANN algorithms so that they cannot be directly applied to SNNs because of the neural dynamic of spiking neurons.
	
	In this paper, we propose a full-stack solution towards faster, larger, and better SNNs from both algorithm and programming aspects. We draw inspirations from the recent work \cite{Wu2018Spatio}, which proposes the spatio-temporal back propagation (STBP) method for the direct training of SNNs, and significantly extend it to much deeper structure, larger dataset, and better performance. We first propose the NeuNorm method to balance neural selectivity and increase the performance. Then we
	improve the rate coding to fasten the convergence and convert the LIF model into an explicitly iterative version to make it compatible with a machine learning framework (Pytorch). As a results, compared to the running time on Matlab, we achieve tens of times speedup that enables the direct training of deep SNNs. The best accuracy on neuromorphic datasets (N-MNIST and DVS-CIFAR10) and comparable accuracy as existing ANNs and pre-trained SNNs on non-spiking datasets (CIFAR10) are demonstrated. This work enables the exploration of direct training of high-performance SNNs and facilitates the SNN applications via compatible programming within the widely used machine learning (ML) framework. 	
	\section{Related work}
	
	We aims to direct training of deep SNNs. To this end, we mainly make improvements on two aspects: (1) learning algorithm design; (2) training acceleration optimization. We chiefly overview the related works in recent years.
	
	\textbf{Learning algorithm for deep SNNs}. There exist three ways for SNN learning: i) unsupervised learning such as spike timing dependent plasticity (STDP); ii) indirect supervised learning such as ANNs-to-SNNs conversion; iii) direct supervised learning such as gradient descent-based back propagation. However, by far most of them limited to very shallow structure (network layer less than 4) or toy small dataset (e.g. MNIST, Iris), and little work points to direct training deep SNNs due to their own challenges.
	
	STDP is biologically plausible but the lack of global information hinders the convergence of large models, especially on complex datasets \cite{Timoth2007Unsupervised,Diehl2015Unsupervised,Tavanaei2017Bio}. 
	
	The ANNs-to-SNNs conversion \cite{Diehl2015Fast,Cao2015Spiking,Sengupta2018Going,Hu2018Spiking} is currently the most successful method to model large-scale SNNs. They first train non-spiking ANNs and convert it into a spiking version. However, this indirect training help little on revealing how SNNs learn since it only implements the inference phase in SNN format (the training is in ANN format). Besides, it adds many constraints onto the pre-trained ANN models, such as no bias term, only average pooling, only ReLU activation function, etc. With the network deepens, such SNNs have to run unacceptable simulation time (100-1000 time steps) to obtain good performance. 
	
	The direct supervised learning method trains SNNs without conversion. They are mainly based on the conventional gradient descent. Different from the previous spatial backpropagation \cite{Lee2016Training,Jin2018Hybrid},   \cite{Wu2018Spatio}  proposed the first backpropagation in both spatial and temporal domains to direct train SNNs, which achieved state-of-the-art accuracy on MNIST and N-MNIST datasets. However, the learning algorithm is not optimized well and the slow simulation hinders the exploration onto deeper structures.
	

	\textbf{Normalization}. Many normalization techniques have been proposed to improve the convergence \cite{Ioffe2015Batch,Wu2018Group}.	Although these methods have achieved great success in ANNs, they are not suitable for SNNs due to the complex neural dynamics and binary spiking representation of SNNs. Furthermore, in terms of hardware implementation, batch-based normalization techniques essentially incur lateral operations across different stimuluses which are not compatible with existing neuromorphic platforms \cite{Merolla2014Artificial,Davies2018Loihi}. Recently, several normalization methods (e.g. model-based normalization \cite{Diehl2015Fast}, data-based normalization \cite{Diehl2015Fast}, spike-based normalization \cite{Sengupta2018Going}) have been proposed to improve SNN performance. However, such methods are specifically designed for the indirect training with ANNs-to-SNNs conversion, which didn't show convincing effectiveness for direct training of SNNs targeted by this work.
	
	\textbf{SNN programming frameworks}. There exist several programming frameworks for SNN modeling, but they have different aims. NEURON \cite{Carnevale2006The} and Genesis \cite{Bower1998The} mainly focus on the biological realistic simulation from neuron functionality to synapse reaction, which are more beneficial for the neuroscience community. BRIAN2 \cite{Goodman2009The} and NEST \cite{Gewaltig2007NEST} target the simulation of larger scale SNNs with many biological features, but not designed for the direct supervised learning for high performance discussed in this work. BindsNET \cite{Hazan2018BindsNET} is the first reported framework towards combining SNNs and practical applications. However, to our knowledge, the support for direct training of deep SNNs is still under development. Furthermore, according to the statistics from ModelDB \footnote{ModelDB is an open  website for storing and sharing computational neuroscience models.},	in most cases researchers even simply program in general-purpose language, such as C/C++ and Matlab, for better flexibility. Besides the different aims, programming from scratch on these frameworks is user unfriendly and the tremendous execution time impedes the development of large-scale SNNs. Instead of developing a new framework, we provide a new solution to establish SNNs by virtue of mature ML frameworks which have demonstrated easy-to-use interface and fast running speed.
	
	\section{Approach}	
	
	In this section, we first convert the LIF neuron model into an easy-to-program version with the format of explicit iteration. Then we propose the NeuNorm method to adjust the neuronal selectivity for improving model performance. Furthermore, we optimize the rate coding scheme from encoding and decoding aspects for faster response. Finally, we describe the whole training and provide the pseudo codes for Pytorch.
	
	\subsection{Explicitly iterative LIF model}
	LIF model is commonly used to describe the behavior of neuronal activities, including the update of membrane potential and spike firing.	Specifically, it is governed by
	
\begin{figure} [!htb]
		\centering
		\includegraphics[width=8.5cm]{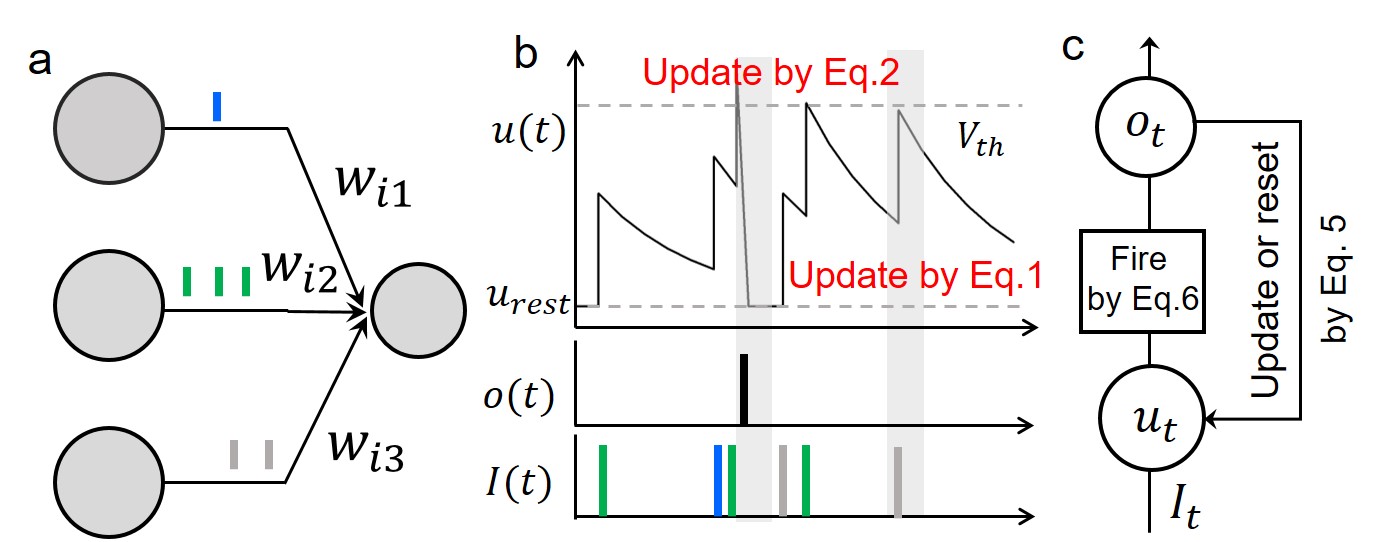}
		\caption{\textbf{Illustration of iterative LIF}. Spike communication between neurons; (b) The update of membrane potential according to Eq. (\ref{LIF_1})-Eq. (\ref{LIF_1_v}); (c) Iterative LIF described by Eq. (\ref{LIF_STD_1})-Eq. (\ref{LIF_STD_2}).}
		\label{LIF}
	\end{figure}

	\begin{footnotesize}
		\begin{align}
		\tau \frac{du}{dt} = -u + I, \quad u<V_{th}
		\label{LIF_1}
		\\
		\text{fire a spike}~\&~u = u_{reset}, \quad u \ge V_{th} 
		\label{LIF_1_v}	
		\end{align}
	\end{footnotesize}	
	where $u$ is the membrane potential, $\tau$ is a time constant, $I$ is pre-synaptic inputs, and $V_{th}$ is a given fire threshold. Eq. (\ref{LIF_1})-(\ref{LIF_1_v}) describe the behaviors of spiking neuron in a way of updating-firing-resetting mechanism (see Fig. \ref{LIF}a-b). When the membrane potential reaches a given threshold, the neuron will fire a spike and $u$ is reset to $u_{reset}$; otherwise, the neuron receives pre-synapse stimulus $I$ and updates its membrane potential according to Eq. (\ref{LIF_1}).

	The above differential expressions described in continuous domain are widely used for biological simulations, but it is too implicit for the implementations on mainstream ML frameworks (e.g. Tensorflow, Pytorch) since the embedded automatic differentiation mechanism executes codes in a discrete and sequential way \cite{paszke2017automatic}. Therefore, it motivates us to convert Eq. (\ref{LIF_1})-(\ref{LIF_1_v}) into an explicitly iterative version for ensuring 
	computational tractability. To this end, we first use Euler method to solve the first-order differential equation of Eq. (\ref{LIF_1}), and obtain an iterative expression
	\begin{footnotesize}
		\begin{align}
		u^{t+1} =  (1-\frac{dt}{\tau}) u^{t} + \frac{dt}{\tau}I.
		\label{LIF_2}
		\end{align}	
	\end{footnotesize}	
	We denote the factor $ (1-\frac{dt}{\tau})$ as a decay factor $k_{\tau 1}$ and expand the pre-synaptic input $I$ to a linearing summation $\sum_j{W_{j}o(j)}$. The subscript $j$ indicates the index of pre-synapse and $o(j)$ denotes the corresponding pre-synaptic spike which is binary (0 or 1). By incorporating the scaling effect of $\frac{dt}{\tau}$ into synapse weights $W$,  we have	
	\begin{footnotesize}
		\begin{align}
		u^{t+1} = k_{\tau 1} u^{t} + \sum_j{W_{j}o(j)}.
		\label{LIF_3}
		\end{align}
	\end{footnotesize}

	Next we add the firing-and-resetting mechanism to Eq. (\ref{LIF_3}). By assuming $u_{reset} = 0$ as usual, we get the final update equations as below
	\begin{footnotesize}
		\begin{align}
		\label{LIF_STD_1}
		u^{t+1,n+1}(i) &= k_{\tau 1}u^{t,n+1}(i)(1 - o^{t,n+1}(i))+\sum_{j=1}^{l(n )}w_{ij}^{n} o^{t+1,n}(j), \\
		\label{LIF_STD_2}
		o^{t+1,n+1}(i) &= f(u^{t+1,n+1}(i) - V_{th}),
		\end{align}
	\end{footnotesize}
	where $n$ and $l(n)$ denote the $n$-th layer and its neuron number, respectively, $w_{ij}$ is the synaptic weight from the $j$-th neuron in pre-layer ($n$) to the $i$-th neuron in the post-layer ($n+1$). $f(\cdot)$ is the step function, which satisfies  $f(x)=0$ when $x<0$, otherwise $f(x)=1$ . Eq (\ref{LIF_STD_1})-(\ref{LIF_STD_2}) reveal that firing activities of $o^{t,n+1}$ will affect the next state $o^{t+1,n+1}$ via the updating-firing-resetting mechanism  (see Fig. \ref{LIF}c). If the neuron emits a spike at time step $t$, the membrane potential at $t+1$ step will clear its decay component $k_{\tau 1} u_{t}$ via the term $(1-o_j^{t,n+1})$, and vice versa. 
	
	Through Eq. (\ref{LIF_STD_1})-(\ref{LIF_STD_2}), we convert the implicit Eq. (\ref{LIF_1})-(\ref{LIF_1_v}) into an explicitly version, which is easier to implement on ML framework. We give a concise pseudo code based on Pytorch for an explicitly iterative LIF model in Algorithm \ref{Algorithm 1}.

	\begin{algorithm}[!htb]
		\caption{ State update for an explicitly iterative LIF neuron at time step $t+1$ in the $(n+1)$-th layer.} 
		\label{Algorithm 1}
		\begin{small}
			\begin{algorithmic}[1] 
				\REQUIRE  Previous potential $u^{t,n+1}$ and spike output $o^{t,n+1}$, current spike input $o^{t+1,n}$, and weight vector $W$
				\ENSURE Next potential $u^{t+1,n+1}$ and spike output $o^{t+1,n+1}$ \\	 
				\textbf{Function} StateUpdate($W$,$u^{t,n+1}, o^{t,n+1}, o^{t+1,n})$\\
				\STATE  \quad $u^{t+1,n+1} = k_{\tau 1} u^{t,n+1}(1-o^{t,n+1})+Wo^{t+1,n }$;\\   
				\STATE \quad $o^{t+1,n+1} = f(u^{t+1,n+1}- V_{th})$ \quad 
				
				\RETURN $u^{t+1,n+1}, o^{t+1,n+1}$
				
			\end{algorithmic}
		\end{small}
	\end{algorithm}	
	\subsection{Neuron normalization (NeuNorm)}	
	Considering the signal communication between two convolutional layers, the neuron at the location $(i, j)$ of the $f$-th feature map (FM) in the $n$-th layer receives convolutional inputs $I$, and updates its membrane potential by
	\begin{footnotesize}
		\begin{align}
		\label{Conv_u}
		u_f^{t+1,n+1}(i,j) &= k_{\tau 1}u_f^{t,n+1}(i,j)(1 -  o_f^{t,n+1}(i,j) )+I_f^{t+1,n+1}(i,j),\\
		\label{Conv_input}
		I_f^{t+1,n+1}(i,j) &= \sum_{c}W_{c,f}^{n}\circledast o^{t+1,n}_c(R{(i,j)}), 
		\end{align}
	\end{footnotesize}
	where $W_{c,f}^{n}$ denotes the weight kernel between the $c$-th FM in layer $n$ and the $f$-th FM in layer $n+1$, $\circledast$ denotes the convolution operation, and $R{(i,j)}$ denotes the local receptive filed of location $(i, j)$.
	\begin{figure} [!htbp]
		
		\centering
		\includegraphics[width=6cm]{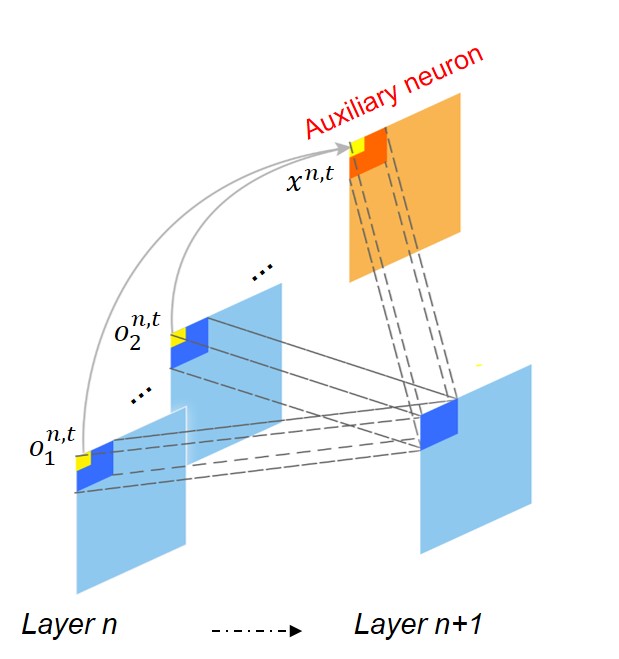}
		\caption{\textbf{Illustration of NeuNorm}. Blue squares represent the normal neuronal FMs, and orange squares represents the auxiliary neuronal FMs. The auxiliary neurons receive lateral inputs (solid lines), and fire signals (dotted lines) to control the strength of stimulus emitted to the next layer.}
		\label{NeuNorm}
		
	\end{figure}
	
	An inevitable problem for training SNNs is to balance the whole firing rate because of the binary spike communication \cite{Diehl2015Fast,Wu2018Spatio}. That is to say, we need to ensure timely and selective response to pre-synaptic stimulus but avoid too many spikes that probably harm the effective information representation. In this sense, it requires that the strength of stimulus maintains in a relatively stable range to avoid activity vanishing or explosion as network deepens. 
	
	Observing that neurons respond to stimulus from different FMs, it motivates us to propose an auxiliary neuron method for normalizing the input strength at the same spatial locations of different FMs (see Fig. \ref{NeuNorm}). The update of auxiliary neuron status $x$ is described by 
	\begin{footnotesize}
		\begin{align}
		\label{Conv_x}
		x^{t+1,n}(i,j)=k_{\tau 2}x^{t,n}(i,j)+ \frac{v}{F} \sum_f  o^{t+1,n}_{f}(i,j),
		\end{align}
	\end{footnotesize}
	where $k_{\tau 2}$ denotes the decay factor, $v$ denotes the constant  scaling factor, and $F$ denotes the number of FMs in the $n$-th layer. In this way, 	 Eq.(\ref{Conv_x}) calculates the average response to the input firing rate with momentum term $k_{\tau 2}x^{t,n}(i,j)$. For simplicity we set $k_{\tau_2}+vF = 1$.
	
	Next we suppose that the auxiliary neuron receives the lateral inputs from the same layer, and transmits signals to control the strength of stimulus emitted to the next layer through trainable weights $U_c^n$, which has the same size as the FM. Hence, the inputs $I$ of neurons in the next layer can be modified by
	\begin{footnotesize}
		\begin{align}
		\label{Modifed_Input}	
		\nonumber
		I_f^{t+1,n+1}(i,j) = \sum_{c}W_{c,f}^{n} \circledast ( o_c^{t+1,n}(R(i,j))-...\\		
		U_{c}^n(R(i,j))\cdot x^{t+1,n}(R(i,j))).	 
		\end{align}
	\end{footnotesize}
	
	Inferring from Eq. (\ref{Modifed_Input}), 	NeuNorm method essentially normalize the neuronal activity by using the input statistics (moving average firing rate). The basic operation is similar with the zero-mean operation in batch normalization. But NeuNorm has different purposes and different data processing ways (normalizing data along the channel dimension rather than the batch dimension).  This difference brings several benefits which may be more suitable for SNNs. Firstly, NeuNorm is compatible with neuron model (LIF) without additional operations and friendly for neuromorphic implementation. Besides that, NeuNorm is more bio-plausible. Indeed, the biological visual pathways are not independent. The response of retina cells for a particular location in an image is normalized across adjacent cell responses in the same layer \cite{Carandini2012Normalization,Mante2008Functional}. And inspired by these founding, many similar techniques for exploiting channel features have been successfully applied to visual recognition, such as group normalization (GN) \cite{Wu2018Group}, DOG \cite{Dalal2005Histograms} and HOG \cite{Dalal2005Histograms}. For example, GN divides FMs of the same layer into several groups and normalizes the features within each group also along channel dimension.
	
	\subsection{Encoding and decoding schemes}
	
	To handle various stimulus patterns, SNNs often use abundant coding methods to process the input stimulus. For visual recognition tasks, a popular coding scheme is rate coding. On the input side, the real-valued images are converted into a spike train whose fire rate is proportional to the pixel intensity. The spike sampling is probabilistic, such as following a Bernoulli distribution or a Poisson distribution. On the output side, it counts the firing rate of each neuron in the last layer over given time windows to determine network output. However, the conventional rate coding suffers from long simulation time to reach good performance. To solve this problem, we take a simpler coding scheme to accelerate the simulation without compromising the performance. 
	
	\textbf{Encoding}. 
	One requirement for long simulation time is to reduce the sampling error when converting real-value inputs to spike signals. Specifically, given the time window $T$, a neuron can represent information by $T+1$ levels of firing rate, i.e. $\{0, \frac{1}{T},...,1\}$ (normalized).  Obviously, the rate coding requires sufficient long window to achieve satisfactory precision. To address this issue, we assign the first layer as an encoding layer and allow it to receive both spike and non-spike signals (compatible with various datasets). In other words, neurons in the encoding layer can process the even-driven spike train from neuromorphic dataset naturally, and can also convert real-valued signals from non-spiking dataset into spike train with enough precision. In this way, the precision is remained to a great extent without depending much on the simulation time. 
	
	\textbf{Decoding}. Another requirement for long time window is the representation precision of network output. To alleviate it, we adopt a voting strategy \cite{Diehl2015Unsupervised} to decode the network output. We configure the last layer as a voting layer consisting of several neuron populations and each output class is represented by one population. 	
	In this way, the burden of representation precision of each neuron in the temporal domain (firing rate in a given time windows) is transferred much to the spatial domain (neuron group coding). Thus, the requirement for long simulation time is significantly reduced. For initialization, we randomly assign a label to each neuron; while during training, the classification result is determined by counting the voting response of all the populations.  
	
	In a nutshell, we reduce the demand for long-term window from above two aspects, which in some sense extends the representation capability of SNNs in both input and output sides. We found similar coding scheme is also leveraged by previous work on ANN compression \cite{tang2017train,Hubara2016Binarized}. It implies that, regardless of the internal lower precision, maintaining the first and last layers in higher precision are important for the convergence and performance. 
	
	\subsection{Overall training implementation}
	
	We define a loss function $L$ measuring the mean square error between the averaged voting results and label vector $Y$ within a given time window $T$ 
	\begin{footnotesize}
		\begin{align}
		\label{loss}
		L =  \parallel   Y  - \frac{1}{T} \sum_{t=1}^T M o^{t,N} \parallel_2^2,
		\end{align}
	\end{footnotesize}
	
	where $o^{t,N}$ denotes the voting vector of the last layer $N$ at time step $t$.  $M$ denotes the constant voting matrix connecting each voting neuron to a specific category.
	
	From the explicitly iterative LIF model, we can see that the spike signals not only propagate through the layer-by-layer spatial domain, but also affect the neuronal states through the temporal domain. Thus, gradient-based training should consider both the derivatives in these two domains. Based on this analysis, we integrate our modified LIF model, coding scheme, and proposed NeuNorm into the STBP method \cite{Wu2018Spatio} to train our network. When calculating the derivative of loss function $L$ with respect to $u$ and $o$ in the $n$-th layer at time step $t$, the STBP propagates the gradients $\frac{\partial L}{\partial o_i^{t,n+1}}$ from the $(n+1)$-th layer and $\frac{\partial L}{\partial o_i^{t+1,n}}$ from time step $t+1$ as follows 
	\begin{footnotesize}
		\begin{align}
		\label{stbp-o}
		\frac{\partial L}{\partial o_i^{t,n}}
		&=
		\sum_{j=1}^{l(n+1)}\frac{\partial L}{\partial o_i^{t,n+1}}\frac{\partial o_j^{t,n+1}}{\partial o_i^{t,n}} + \frac{\partial L}{\partial o_i^{t+1,n}}\frac{\partial o_i^{t+1,n}}{\partial o_i^{t,n}} ,
		\\
		\label{stbp-u}
		\frac{\partial L}{\partial u_i^{t,n}}
		&=
		\frac{\partial L}{\partial o_i^{t,n}} \frac{\partial o_i^{t,n}}{\partial u_i^{t,n}}+\frac{\partial L}{\partial o_i^{t+1,n}}\frac{\partial o_i^{t+1,n}}{\partial u_i^{t,n}} .
		\end{align}
	\end{footnotesize}
	
	\begin{algorithm}[!ht]
		\begin{threeparttable}
			\caption{ Training codes for one iteration.}
			\label{Algorithm 2}
			\begin{small}	
				\begin{algorithmic}[1] 
					\REQUIRE  Network inputs $\{X^t\}_t^T$, class label $Y$, parameters and states of convolutional layers  ($\{W^{l}$, $U^{l}$ , $u^{0,l}$, $o^{0,l}\}_{l=1}^{N_1}$) and fully-connected layers  ($\{W^{l}$, $u^{0,l}$, $o^{0,l}\}_{l=1}^{N_2}$), simulation windows $T$,   voting matrix $M$\quad // Map each voting neuron to label.
					\ENSURE Update network parameters.\\
					~\\					
					\textbf{Forward (inference):} \\
					\FOR{$t = 1$ to $T$}
					\STATE $o^{t,1}$  $\leftarrow$  EncodingLayer($X^t$) \quad

					\FOR{  $l = 2$ to $N_1 $ }
					\STATE  $x^{t,l-1}$  $\leftarrow$  AuxiliaryUpdate($x^{t-1,l-1}$, $o^{t,l-1}$)\quad // Eq. (\ref{Conv_x}).
					\STATE  ($u^{t,l }$, $o^{t,l}$) $\leftarrow$ StateUpdate($W^{l-1}$, $u^{t-1,l }$,  $o^{t-1,l }$, $o^{t,l-1},U^{l-1}, x^{t,l-1 }$)\quad //  Eq. (\ref{LIF_STD_2})-(\ref{Conv_u}), and (\ref{Conv_x})-(\ref{Modifed_Input})	 		
					\ENDFOR
					
					\FOR{$l = 1$ to $N_2$}
					\STATE ($u^{t,l}$, $o^{t,l}$) $\leftarrow$ StateUpdate($W^{l-1}$, $u^{t-1,l}$, $o^{t-1,l}$,  $o^{t,l-1}$)\\ // Eq. (\ref{LIF_STD_1})-(\ref{LIF_STD_2})
					
					\ENDFOR	
					\ENDFOR	\\		
					~\\					
					\textbf{Loss:} \\
					\STATE $L$ $\leftarrow$ ComputeLoss($Y$, $M$, $o^{t,N_2}$)\quad //  Eq. (\ref{loss}). \\				
					~\\
					\textbf{Backward:} \\
					\STATE Gradient initialization: $\frac{\partial L}{\partial o^{T+1,*}} = 0$. 
					\FOR{ $t = T$ to 1 }
					\STATE  $\frac{\partial L}{\partial o^{t,N_2}}$   $\leftarrow$ LossGradient($L,M,\frac{\partial L}{\partial o^{t+1,N_2}}$)\\ // Eq. (\ref{LIF_STD_1})-(\ref{LIF_STD_2}), and  (\ref{loss})-(\ref{stbp-u}).  
					
					\FOR{   $l = N_2-1$ to 1 }
					\STATE ($\frac{\partial L}{\partial o^{t,l}}$, $\frac{\partial L}{\partial u^{t,l}}$, $\frac{\partial L}{\partial W^{l}}$) $\leftarrow$ BackwardGradient($\frac{\partial L}{\partial o^{t,l+1}},$ $\frac{\partial L}{\partial o^{t+1,l}}, W^l$)\quad // Eq. (\ref{LIF_STD_1})-(\ref{LIF_STD_2}),  and (\ref{stbp-o})-(\ref{stbp-u}). \\
					\ENDFOR
					
					\FOR{  $l = N_1$  to 2}
					\STATE ($\frac{\partial L}{\partial o^{t,l}}$, $\frac{\partial L}{\partial u^{t,l}}$, $\frac{\partial L}{\partial W^{l-1}}$, $\frac{\partial L}{\partial U^{l-1}}$) $\leftarrow$ BackwardGradient( $\frac{\partial L}{\partial o^{t,l+1}}$,$\frac{\partial L}{\partial o^{t+1,l}},  W^{l-1},U^{l-1}$)\\ // Eq. (\ref{LIF_STD_2})-(\ref{Conv_u}), (\ref{Conv_x})-(\ref{Modifed_Input}), and (\ref{stbp-o})-(\ref{stbp-u}).
					\ENDFOR	\\
					
					\ENDFOR	\\
					~\\
					\STATE \textbf{Update parameters based on gradients}.	\\
					~\\
					\footnotesize Note: All the parameters and states with layer index $l$ in the $N_1$ or $N_2$ loop belong to the convolutional or fully-connected layers, respectively. For clarity, we just use the symbol of $l$.
				\end{algorithmic}
			\end{small}	
		\end{threeparttable}
		
	\end{algorithm}
	
	A critical problem in training SNNs is the non-differentiable property of the binary spike activities. To make its gradient available, we take the rectangular function $h(u)$ to approximate the derivative of spike activity \cite{Wu2018Spatio}. It yields  
	\begin{footnotesize}
		\begin{align}
		\label{Approximation}
		h(u) = \frac{1}{a} sign(|u-V_{th}|<\frac{a}{2}),
		\end{align}
	\end{footnotesize}
	where the width parameter $a$ determines the shape of $h(u)$. Theoretically, it can be easily proven that Eq. (\ref{Approximation}) satisfies 
	\begin{footnotesize}
		\begin{align}
		\lim \limits_{a\rightarrow0^{+}}h(u)=\frac{df}{du}. 
		\end{align}
	\end{footnotesize}
	Based on above methods, we also give a pseudo code for implementing the overall training of proposed SNNs in Pytorch, as shown in Algorithm \ref{Algorithm 2}.

	\section{Experiment}
	
	We test the proposed model and learning algorithm on both neuromorphic datasets (N-MNIST and DVS-CIFAR10) and non-spiking datasets (CIFAR10) from two aspects: (1) training acceleration; (2) application accuracy. The dataset introduction,  pre-processing, training detail, and parameter configuration are summarized in Appendix. 
	
	%
	%
	
	\subsection{Network structure}
	Tab. \ref{Table-structure1} and Tab. \ref{Table-structure2} provide the network structures for acceleration analysis and accuracy evaluation, respectively. The structure illustrations of what we call AlexNet and CIFARNet are also shown in Appendix, which are for fair comparison to pre-trained SNN works with similar structure and to demonstrate comparable performance with ANNs, respectively. It is worth emphasizing that previous work on direct training of SNNs demonstrated only shallow structures (usually 2-4 layers), while our work for the first time can implement a direct and effective learning for larger-scale SNNs (e.g. 8 layers).
	
	\begin{table}[!htbp]
		\centering
		\caption{Network structures used for training acceleration.}
		\label{Table-structure1}
		
		\scalebox{0.85}{
			\begin{tabular}{ll}
				\hline
				\multicolumn{2}{c}{\large Neuromorphic  Dataset}                       \\ \hline
				Small    & 128C3(Encoding)-AP2-128C3-AP2-512FC-Voting                      \\
				Middle   & 128C3(Encoding)-128C3-AP2-256-AP2-1024FC-Voting                 \\
				Large    & 128C3(Encoding)-128C3-AP2-384C3-384C3-AP2-\\
				& 1024FC-512FC-Voting     \\ \hline
				\multicolumn{2}{c}{\large Non-spiking Dataset}                         \\ \hline
				Small    & 128C3(Encoding)-AP2-256C3-AP2-256FC-Voting                      \\
				Middle   & 128C3(Encoding)-AP2-256C3-512C3-AP2-512FC-Voting                \\
				Large    & 128C3(Encoding)-256C3-AP2-512C3-AP2-1024C3-512C3-\\
				& 1024FC-512FC-Voting \\ \hline				
		\end{tabular}}
	\end{table}

	\begin{table}[!htbp]
		\centering
		\caption{Network structures used for accuracy evaluation.}
		\label{Table-structure2}
		\scalebox{0.85}{
			\begin{tabular}{ll}
				\hline
				\multicolumn{2}{c}{\large Neuromorphic  Dataset}                                                                               \\ \hline
				Our model  & 128C3(Encoding)-128C3-AP2-128C3-256C3-AP2-\\
				& 1024FC-Voting                \\ \hline
				
				\multicolumn{2}{c}{\large Non-spiking Dataset}                                                                        \\ \hline
				AlexNet &  96C3(Encoding)-256C3-AP2-384C3-AP2-384C3-\\
				& 256C3-1024FC-1024FC-Voting                                                       \\
				CIFARNet  & 128C3(Encoding)-256C3-AP2-512C3-AP2-1024C3-\\
				&512C3-1024FC-512FC-Voting \\ \hline				
		\end{tabular}}
		
	\end{table}

	\subsection{Training acceleration}
	
	\textbf{Runtime}. Since Matlab is a high-level language widely used in SNN community, we adopt it for the comparison with our Pytorch implementation. For fairness, we made several configuration restrictions, such as software version, parameter setting, etc. More details can be found in Appendix.
	
	Fig. \ref{runtime} shows the comparisons about average runtime per epoch, where batch size of 20 is used for simulation. Pytorch is able to provide tens of times acceleration on all three datasets. This improvement may be attributed to the specialized optimizations for the convolution operation in Pytorch. In contrast, currently these optimizations
	have not been well supported in most of existing SNN platforms. 
	Hence building spiking model may benefit a lot from DL platform.
	
	\begin{figure}[!htpp]
		\centering
		\includegraphics[width=0.48\textwidth]{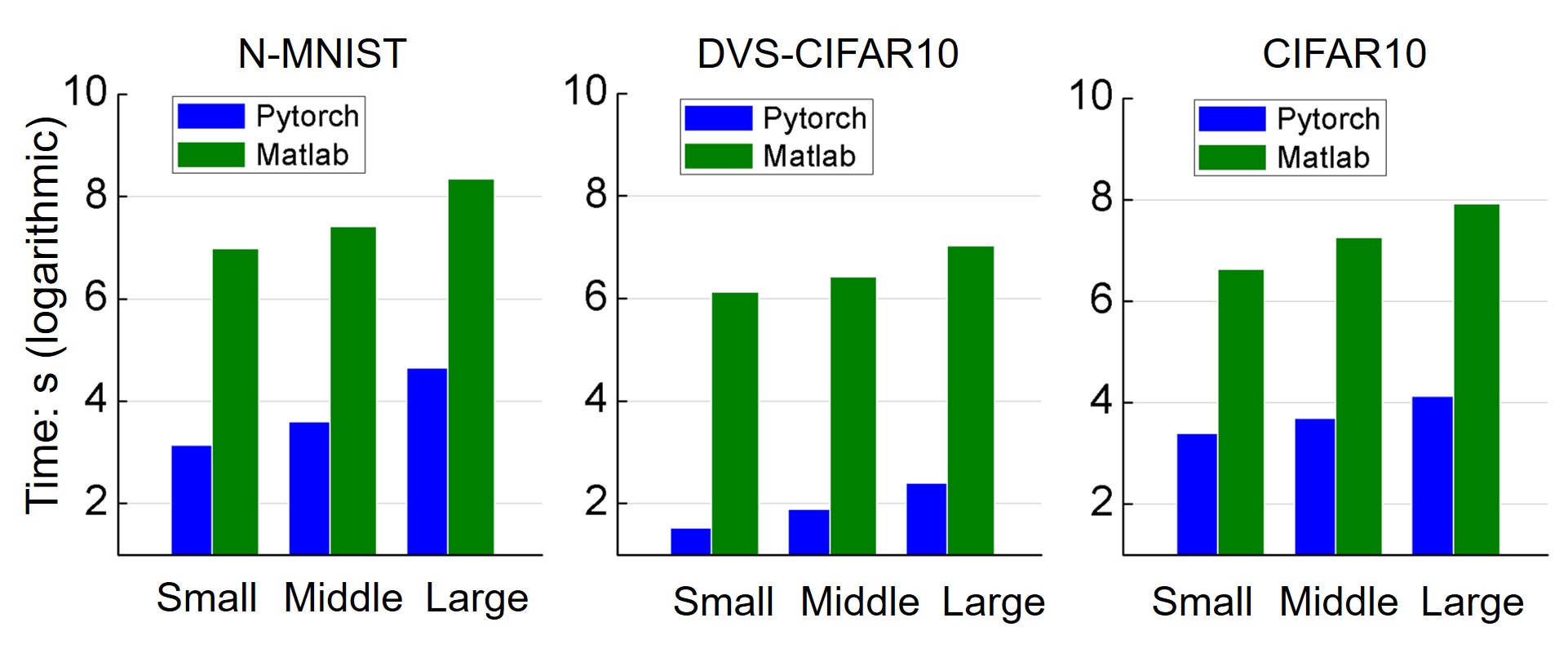}
		\caption{{Average runtime per epoch}.}
		\label{runtime}
		
	\end{figure}
	
\begin{figure}[!htpp] 
		\centering
		\includegraphics[width=0.45\textwidth]{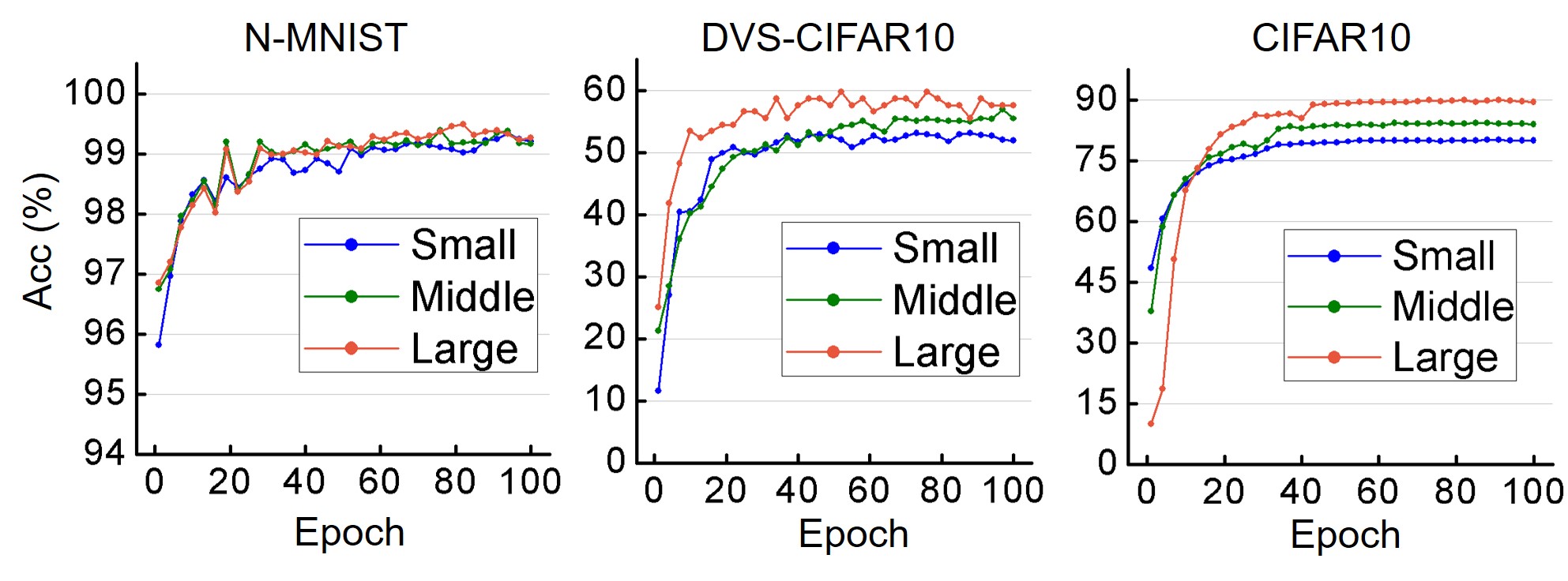}
		\caption{\textbf{Influence of network scale}. Three different network scales are defined in Tab.\ref{Table-structure1}.}
		\label{Net_scale}
		
	\end{figure}

	\textbf{Network scale}. If without acceleration, it is difficult to extend most existing works to large scale. Fortunately, the fast implementation in Pytorch facilitates the construction of deep SNNs. This makes it possible to investigate the influence of network size on model performance. Although it has been widely studied in ANN field, it has yet to be demonstrated on SNNs. To this end, we compare the accuracy under different network sizes, shown in Fig. \ref{Net_scale}. With the size increases, SNNs show an apparent tendency of accuracy improvement, which is consistent with ANNs.
	
\begin{figure} [!htbp]
		\centering
		\includegraphics[width=0.48\textwidth]{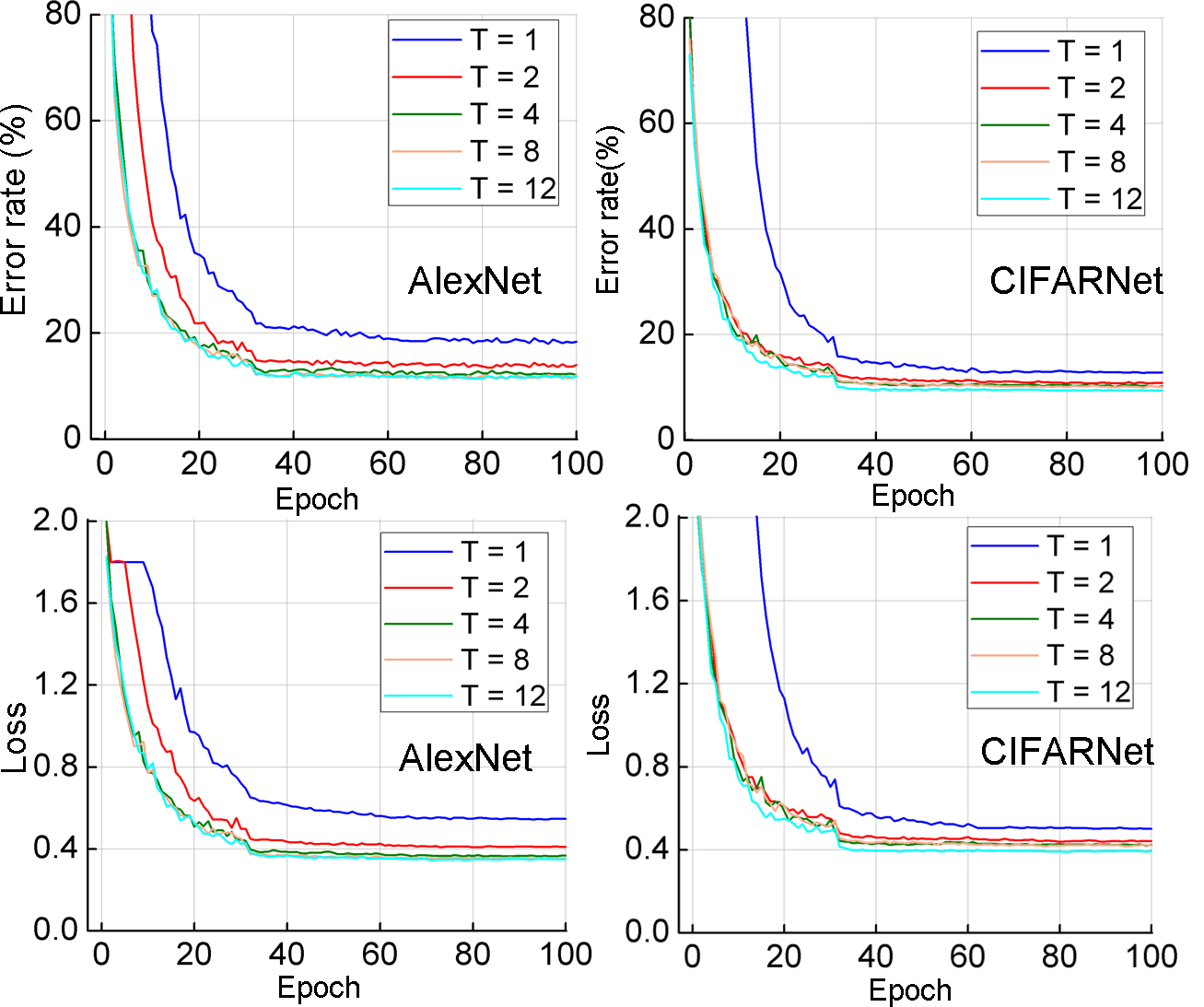}
		\caption{ {Influence of simulation length on CIFAR10.}  } 
		\label{T_length}
		
	\end{figure}

	\textbf{Simulation length}. SNNs need enough simulation steps to mimic neural dynamics and encode information. Given simulation length $T$, it means that we need to repeat the inference process $T$ times to count the firing rate. So the network computation cost can be denoted as $O(T)$. For deep SNNs, previous work usually requires 100 even 1000 steps to reach good performance \cite{Sengupta2018Going}, which brings huge computational cost. Fortunately, by using the proposed coding scheme in this work, the simulation length can be significantly reduced without much accuracy degradation. As shown in Fig. \ref{T_length}, although the longer the simulation windows leads to better results, our method just requires a small length ($4-8$) to achieve satisfactory results. Notably, even if extremely taking one-step simulation, it can also achieve not bad performance with significantly faster response and lower energy, which promises the application scenarios with extreme restrictions on response time and energy consumption.
	
	\subsection{Application accuracy}
	
	\textbf{Neuromorphic datasets}. Tab. \ref{Table-NMNIST} records the current state-of-the-art results on N-MNIST datasets. \cite{Cai2018} proposes a technique to restore the N-MNIST dataset back to the static MNIST, and achieves 99.23\% accuracy using non-spiking CNNs. Our model can naturally handle raw spike streams using direct training and achieves significantly better accuracy. Furthermore, by adding NeuNorm technique, the accuracy can be improved up to 99.53\%. 
	
	\begin{table}[!htbp]
		\centering
		\caption{Comparison with existing results on N-MNIST.}
		\label{Table-NMNIST}
		\scalebox{0.82}{
			\begin{tabular}{ccc}
				\hline
				Model                                                                 & Method                                                                  & Accuracy \\ \hline
				
				\cite{Neil2016Phased}                                                   & LSTM                                                                                   & 97.38\%       \\
				
				\cite{Lee2016Training}                                                 &Spiking NN                                                                 & 98.74\%       \\
				\cite{Wu2018Spatio}                                 &  Spiking NN                                                                 & 98.78\%       \\
				\cite{Jin2018Hybrid}                                  & Spiking NN                                                                  & 98.93\%       \\
				\cite{Neil2016Effective}                                             & Non-spiking CNN                                                                                 & 98.30\%       \\
				\cite{Cai2018}                               & 	Non-spiking CNN                                                                                  & 99.23\%       \\ \hline
				\begin{tabular}[c]{@{}c@{}} \large \textbf{Our model}   \end{tabular} & without NeuNorm & \large \textbf{99.44\%}       \\
				\begin{tabular}[c]{@{}c@{}}\large \textbf{Our model} \end{tabular}    & with NeuNorm & \large \textbf{99.53\%}       \\ \hline
		\end{tabular}}
		
	\end{table}

	\begin{table}[!htbp]
		\centering
		\caption{Comparison with existing results on DVS-CIFAR10.}
		\label{Table-DVSCIFAR10}
		\scalebox{0.9 }{
			\begin{tabular}{ccc}
				\hline
				Model                     & Method            & Accuracy \\ \hline
				\cite{Orchard2015HFirst} &  Random Forest       & 31.0\%   \\
				\cite{Lagorce2017HOTS}           & HOTS       & 27.1\%   \\
				\cite{Sironi2018HATS}               & HAT       & 52.4\%   \\ 
				\cite{Sironi2018HATS}          & Gabor-SNN & 24.5\%   \\ 
				\hline
				\large \textbf{Our model}        & without NeuNorm           & \large \textbf{58.1\%}   \\
				\large \textbf{Our model}           & with NeuNorm           & \large \textbf{60.5\%}   \\ \hline
		\end{tabular}}
		
	\end{table}

	Tab. \ref{Table-DVSCIFAR10} further compares the accuracy on DVS-CIFAR10 dataset. DVS-CIFAR10 is more challenging than N-MNIST due to the larger scale, and it is also more challenging than non-spiking CIFAR10 due to less samples and noisy environment (Dataset introduction in Appendix). Our model achieves the best performance with 60.5\%. Moreover, experimental results indicate
	that NeuNorm can speed up the convergence.Without
	NeuNorm the required training epochs to get the best accuracy
	58.1\% is 157, while it’s reduced to 103  with NeuNorm.
	
	\textbf{Non-spiking dataset}. Tab. \ref{Table-cifar10} summarizes the results of existing state-of-the-art results and our CIFARNet on CIFAR10 dataset. Prior to our work, the best direct training method only achieves 75.42\% accuracy. Our model is significantly better than this result, with 15\% improvement (reaching 90.53\%). We also make a comparison with non-spiking ANN model and other pre-trained works (ANNs-to-SNNs conversion) on the similar structure of AlexNet, wherein our direct training of SNNs with NeuNorm achieves slightly better accuracy. 
	
	\begin{table} [!htbp]
		\centering
		\caption{Comparison with existing state-of-the-art results on non-spiking CIFAR10.}
		\label{Table-cifar10}
		\scalebox{0.9 }{
			\begin{tabular}{ccc}
				\hline
				Model                              & Method            & Accuracy \\ \hline
				\cite{Panda2016Unsupervised}            & Spiking NN   & 75.42\%  \\
				\cite{Cao2015Spiking}              & Pre-trained SNN & 77.43\%  \\
				
				\cite{Rueckauer2017Conversion} & Pre-trained SNN & 90.8\%    \\
				\cite{Sengupta2018Going}                 & Pre-trained SNN & 87.46\%  \\
				\hline
				\large Baseline               & Non-spiking NN                & \large \textbf{90.49\%}   \\
				\large \textbf{Our model}       & without NeuNorm   & \large \textbf{89.83\%}   \\
				\large \textbf{Our model}          & with NeuNorm   & \large \textbf{90.53\%}  \\ \hline
		\end{tabular}}	
		
	\end{table}
	
	\begin{table}[!htbp]
		\centering
		\caption{Comparison with previous results using similar AlexNet struture on non-spiking CIFAR10.}
		\label{Table2-cifar10}
		\scalebox{0.9}{
			\begin{tabular}{ccc}
				\hline
				Model                     & Method                   & Accuracy         \\ \hline
				\cite{Hunsberger2015Spiking} & Non-spiking NN          & 83.72\%          \\
				\cite{Hunsberger2015Spiking}     & Pre-trained SNN              & 83.52\%          \\
				\cite{Sengupta2018Going}      & Pre-trained SNN    & 83.54\%          \\ \hline
				\large \textbf{Our model}  &  ---  &  \large \textbf{85.24\%} \\ \hline
		\end{tabular}}
	\end{table}
	
	\begin{figure} [!ht]
		\centering
		\includegraphics[width=8.0cm]{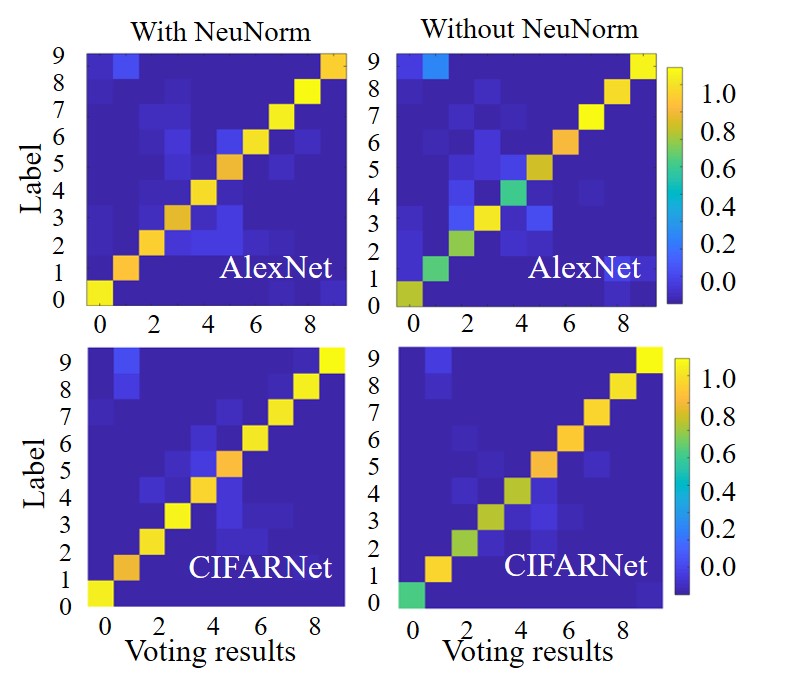}
		\caption{\textbf{Confusion matrix of voting output with or without NeuNorm}. High values along the diagonal indicate correct recognition whereas high values anywhere else indicate confusion between two categories.}
		\label{Confusion_matrix}
		
	\end{figure}

	Furthermore, to visualize the differences of learning with or without NeuNorm, Fig. \ref{Confusion_matrix} shows the average confusion matrix of network voting results on CIFAR10 over 500 randomly selected images. Each location in the 10$\times$10 tiles is determined by the voting output and the actual labels. High values along the diagonal indicate correct recognition whereas high values anywhere else indicate confusion between two categories. It shows that using NeuNorm brings a clearer contrast (i.e. higher values along the diagonal), which implies that NeuNorm enhances the differentiation degree among the 10 classes. Combining the results from Tab. \ref{Table-NMNIST}-\ref{Table-cifar10}, it also confirms the effectiveness of proposed NeuNorm for performance improvement.

	\section{Conclusion}

	In this paper, we present a direct training algorithm for  {deeper and larger} SNNs with high performance. We propose the NeuNorm to effectively normalize the neuronal activities and improve the performance. Besides that, we optimize the rate coding from encoding aspect and decoding aspect, and convert the original continuous LIF model to an explicitly iterative version for friendly Pytorch implementation. Finally, through tens of times training accelerations and larger network scale, we achieve the best accuracy on neuromorphic datasets and comparable accuracy with ANNs on non-spiking datasets. To our best knowledge, this is the first time report such high performance with direct training on SNNs. The implementation on mainstream ML framework could facilitate the SNN development.
	
		\section*{Acknowledgments} 
	The work was supported by the Project of NSFC (No. 61836004, 61620106010, 61621136008, 61332007, 61327902, 61876215), the National Key Research and Development Program of China (No. 2017YFA0700900), the Brain-Science Special Program of Beijing (Grant Z181100001518006), the Suzhou-Tsinghua innovation leading program 2016SZ0102, and the Project of NSF (No. 1725447, 1730309).

	\selectfont
	\bibliographystyle{AAAI}
	\bibliography{bib_test}
	
	\begin{appendices}
		\section{Supplementary material}
		In this supplementary material, we provide the details of our experiment, including the dataset introduction, training setting, and programing platform comparison.
		\section{A $\quad$ Dataset Introduction}
		\subsection{Neuromorphic dataset}
		\textbf{N-MNIST} converts the frame-based MNIST handwritten digit dataset into its DVS (dynamic vision sensor) version \cite{Orchard2015Converting} with event streams (in Fig.\ref{fig2}). For each sample, DVS scans the static MNIST image along given directions and collects the generated spike train which is triggered by detecting the intensity change of pixels. Since the intensity change has two directions (increase or decrease), DVS can capture two kinds of spike events, denoted as On-event and Off-event. Because of the relative shift of images during moving process, the pixel dimension is expanded to 34$\times$34. Overall, each sample in N-MNIST is a spatio-temporal spike pattern with size of $34 \times 34 \times2 \times T$,  where $T$ is the length of temporal window. 
		
		\textbf{DVS-CIFAR10} converts 10000 static CIFAR10 images into the format of spike trains (in Fig.\ref{fig2}), consisting of 1000 images per class with size of $128\times128$ for each image  . Since different DVS types and different movement paths are used \cite{Li2017CIFAR10}, the generated spike train contains imbalanced spike events and larger image resolution. We adopt different parameter  configurations, shown in Tab. \ref{Table-parameters}.  DVS-CIFAR10 has 6 times less samples than the original CIFAR10 dataset, and we randomly choose 9000 images for training and 1000 for testing.
		
		\begin{figure}[!hpt] 
			\centering		 
			\includegraphics[width=7.0cm]{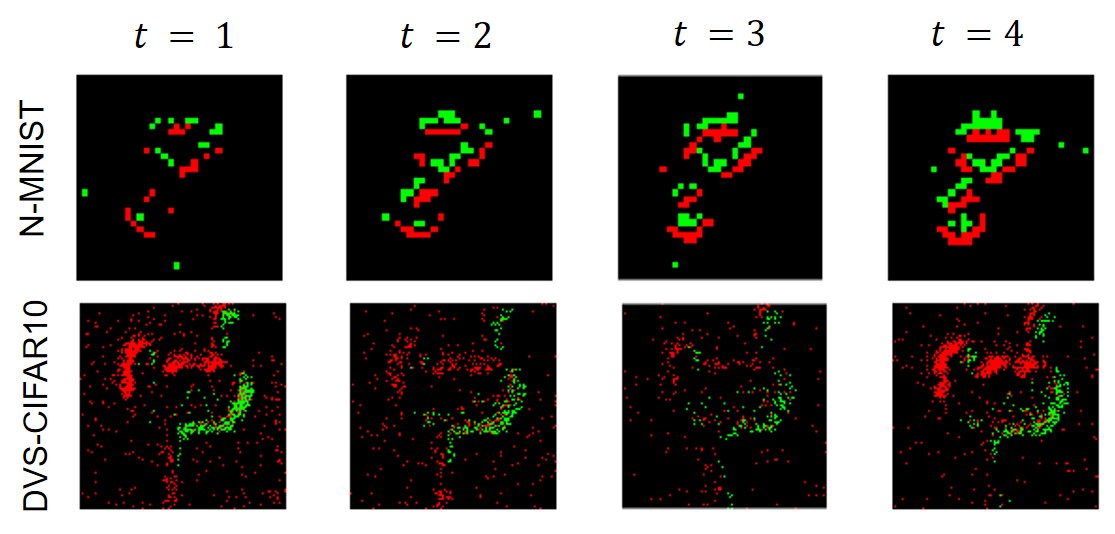}
			\caption{\textbf{Illustration of neuromorphic datasets}. The upper $32 \times 32 $ image and the lower $128 \times 128 $ image are sampled from N-MNIST and DVS-CIFAR10, respectively. Each sub-picture shows a 5 ms-width spike train.}
			\label{Spike_dataset}
			
		\end{figure}
		\subsection{Non-spiking dataset}
		\textbf{CIFAR10} is widely used in ANN domain, which contains 60000 color images belonging to 10 classes with size of $32\times32$ for each image. We divide it into 50000 training images and 10000 test images as usual.

		\section{B $\quad$ Training setting}
		
		\subsubsection{Data pre-processing}
		On N-MNIST, we reduce the time resolution by accumulating the spike train within every 5 ms. On DVS-CIFAR10, we reduce the spatial resolution by down-sampling the original 128 $\times$128 size to 42$\times$42 size (stride = 3, padding = 0), and reduce the temporal resolution by similarly accumulating the spike train within every 5 ms. On CIFAR10, as usual, we first crop and flip the original images along each RGB channel, and then  rescale each image by subtracting the global mean value of pixel intensity and dividing by the resulting standard variance along each RGB channel.

		\subsubsection{Optimizer}On neuromorphic datasets, we use Adam (adaptive moment estimation) optimizer. On the non-spiking dataset, we use the stochastic gradient descent (SGD) optimizer with initial learning rate $r=0.1$ and momentum 0.9, and we let $r$ decay to  $0.1r$ over each 40 epochs. 
		\begin{figure} [!htbp]
			\centering
			\includegraphics[width=8.0cm]{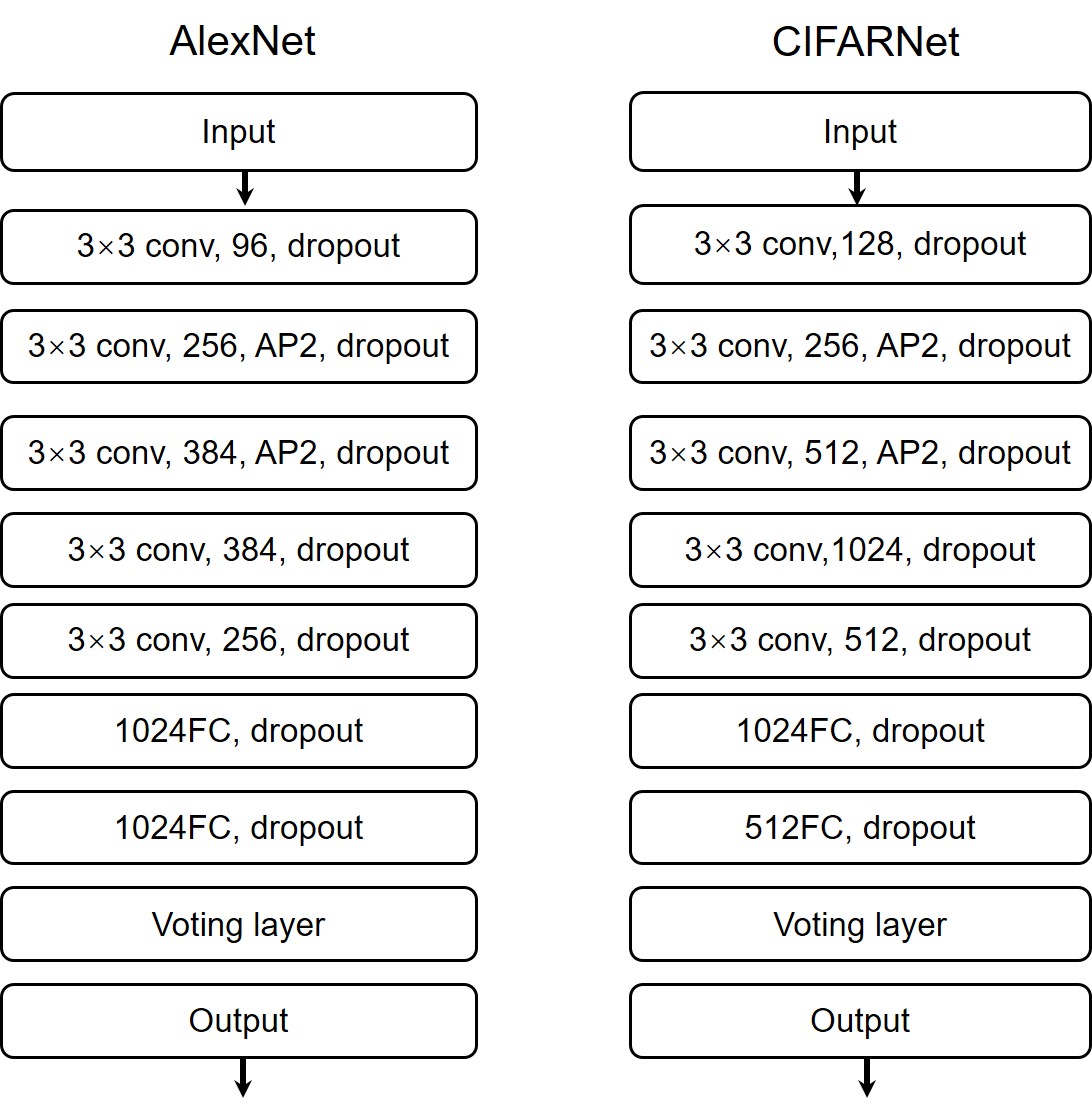}
			\caption{\textbf{Illustration of network structures}. For simplicity, we denote the left as AlexNet and the right as CIFARNet.}
			\label{fig2}
		\end{figure}
		
		\begin{table}[!htb]
			\centering
			\caption{Parameter configuration used for model evaluation.}
			\label{Table-parameters}
			\scalebox{1}{
				\begin{tabular}{cccc}
					\hline
					Parameters & CIFAR10 & DVS-CIFAR10 & N-MNIST \\ \hline
					$V_{th}$ & 0.75 & 0.05 & 0.25 \\
					$a$ & 1.0 & 0.1 & 0.25 \\

					$k_{\tau 1}$ & 0.25 & 0.35 & 0.3 \\
					$k_{\tau 2}$ & 0.9 & 0.9 & 0.9 \\
					Dropout rate & 0.5 & 0 & 0 \\
					Max epoch & 150 & 200 & 200 \\ \hline
					Adam & \multicolumn{3}{c}{$\lambda, \beta_1,\beta_2$ = 0.9,0.999, 1-$10^{-8}$} \\ \hline
			\end{tabular}}
		\end{table}
		\subsubsection{Parameter configuration}The configuration of simulation parameters for each dataset is shown in Tab. \ref{Table-parameters}. The structures of what we call AlexNet and CIFARNet are illustrated in Fig. \ref{fig2}.

		\section{C $\quad$ Details for Acceleration experiment}
		A total of 10 experiments are counted for average.  Considering the slow running time on the Matlab, 10 simulation steps and batch size of 20  per epoch are used. All codes run on server with i7-6700K CPU and GTX1060 GPU. The Pytorch version is 3.5 and the Matlab version is 2018b. Both of them enable GPU execution, but without parallelization. 
		
		It is worth noting that recently there emerge more supports on Matlab for deep learning, either official \footnote{Since 2017 version, Matlab provides deep learning libraries.} or unofficial (e.g. MatConvNet toolbox\footnote{A MATLAB toolbox designed for implementing CNNs. }). However, their highly encapsulated function modules are unfriendly for users to customize deep SNNs. Therefore, in terms of flexibility, all convolution operations we adopted are based on the officially provided operations (i.e. $conv$ function).
	\end{appendices}

\end{document}